\documentclass{article}

\usepackage{arxiv}

\usepackage[utf8]{inputenc} 
\usepackage[T1]{fontenc}    
\usepackage{hyperref}       
\usepackage{url}            
\usepackage{booktabs}       
\usepackage{amsfonts}       
\usepackage{nicefrac}       
\usepackage{microtype}      
\usepackage{lipsum}     
\usepackage{graphicx}
\usepackage{natbib}
\usepackage{doi}

\title{Learn to Learn Metric Space for Few-Shot Segmentation of 3D Shapes}

\date{} 

\renewcommand{\headeright}{Li et al.}
\renewcommand{\undertitle}{}
\renewcommand{\shorttitle}{\textit{arXiv}}

\author{
    Xiang Li \\
    New York University Abu Dhabi \\
    \texttt{xl1845@nyu.edu} \\
    \And
    Lingjing Wang \\
    New York University Abu Dhabi \\
    \texttt{lingjing.wang@nyu.edu} \\
    \And
    Yi Fang\thanks{Corresponding author} \\
    New York University Abu Dhabi \\
    \texttt{yfang@nyu.edu} \\
}

\begin{document}

\maketitle

\begin{abstract}
Recent research has seen numerous supervised learning based methods for 3D shape segmentation and remarkable performance has been achieved on various benchmark datasets. These supervised methods require a large amount of annotated data to train deep neural networks to ensure the generalization ability on the unseen test set. In this paper, we introduce a meta-learning-based method for few-shot 3D shape segmentation where only a few labeled samples are provided for the unseen classes. To achieve this, we treat the shape segmentation as a point labeling problem in the metric space. Specifically, we first design a meta-metric learner to transform input shapes into embedding space and our model learns to learn a proper metric space for each object class based on point embeddings. Then, for each class, we design a metric learner to extract part-specific prototype representations from a few support shapes and our model performs per-point segmentation over the query shapes by matching each point to its nearest prototype in the learned metric space. A metric-based loss function is used to dynamically modify distances between point embeddings thus maximizes in-part similarity while minimizing inter-part similarity. A dual segmentation branch is adopted to make full use of the support information and implicitly encourages consistency between the support and query prototypes. We demonstrate the superior performance of our proposed on the ShapeNet part dataset under the few-shot scenario, compared with well-established baseline and state-of-the-art semi-supervised methods.
\end{abstract}


\section{Introduction}
3D shape segmentation is generally defined as identifying the semantic part category of each point in the input 3D shape. Recent years, with the prosperous of deep learning-based methods for various vision-related applications, researchers have developed numerous deep learning-based methods for 3D tasks, such as object classification \citep{qi2017pointnet}, part segmentation \citep{li2018so}, point cloud segmentation \citep{thomas2019kpconv}, shape registration \citep{wang2019deep} and shape correspondence \citep{litany2017deep}. Unlike traditional shallow models that rely on hand-crafted features, deep learning-based methods can learn more representative features through multiple layers and thus can better resolve input ambiguities (e.g. object size, orientations, structural variations, incompleteness, and noise) and produce robust segmentation of 3D shapes \citep{qi2017pointnet,su2018splatnet,li2018pointcnn}. 

Existing deep learning-based methods for 3D shape segmentation can be roughly divided into two categories. The first type of method starts with converting irregular point clouds into regular representations, such as 2D images or 3D voxels \citep{su2015multi,wang2017dominant,zhou2018voxelnet,wu2019pointconv}, and then apply conventional 2D/3D CNN to conduct segmentation. Another family of methods directly apply convolution on raw point clouds to learn point embeddings, followed by the fully connected layers to perform per-point classification. PointNet and its variants \citep{qi2017pointnet++,li2018so,jiang2018pointsift} have been showing ever-increasing performance for 3D shape segmentation on various benchmark datasets. Nevertheless, these methods mostly train deep neural networks in a supervised manner and thus require a huge number of labeled data to learn the network parameters. However, it is practically labor-intensive and time-consuming to collect large-scale annotated datasets with dense semantic part labels, which impedes the application of learning-based methods in broader 3D segmentation applications.

\begin{figure*}[t]
\begin{center}
\includegraphics[width=16cm,]{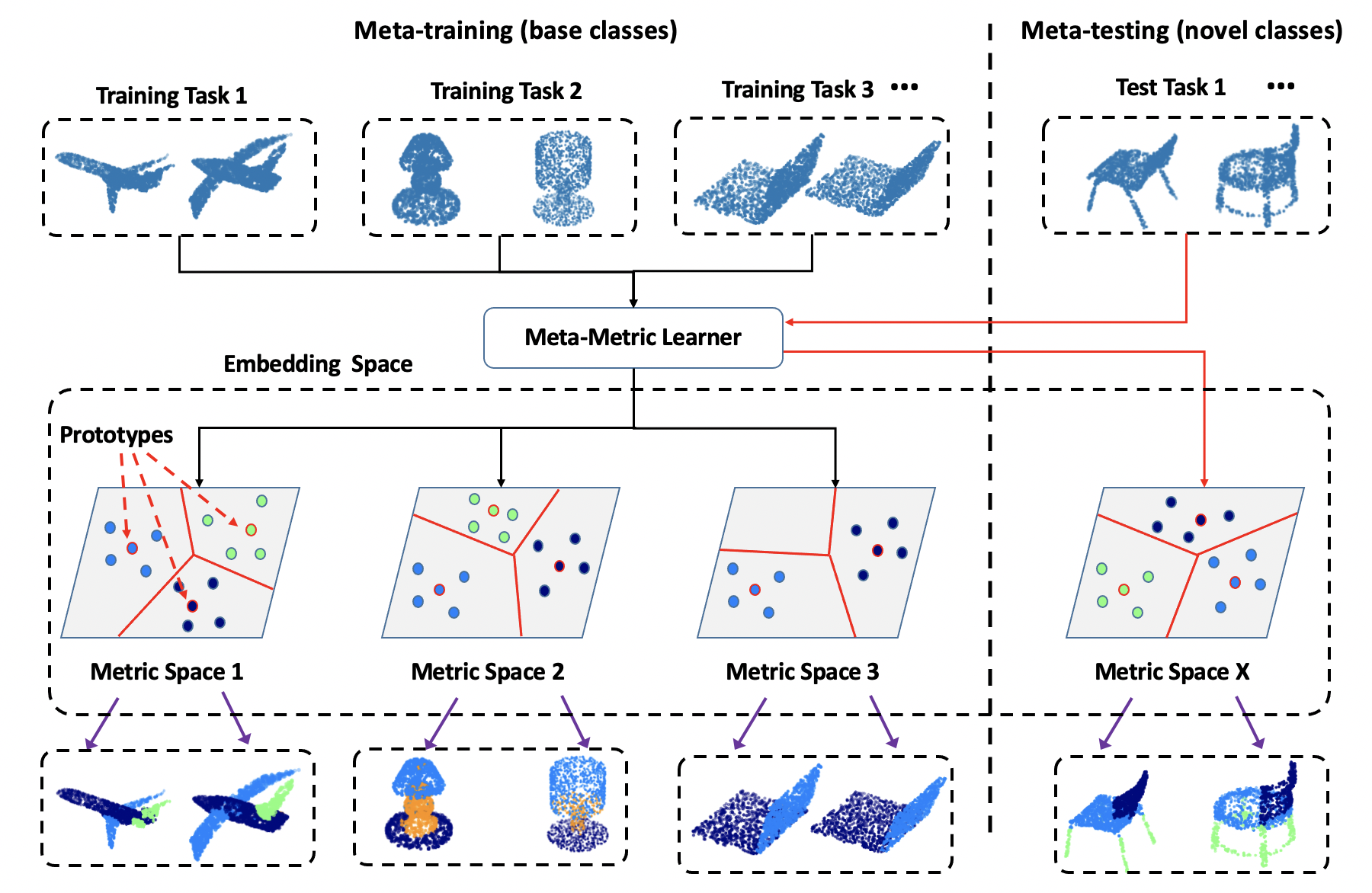}
\end{center}
\caption{Illustration of meta-learning-based part segmentation. 
Our model starts by a meta-metric learner that transforms input shapes into embedding space where the shape segmentation metrics are defined. Then, for each learning task, our method learns to learn a metric space where points from the same semantic part category cluster around a single prototype representation (points with red outlines), and part segmentation labels can be assigned by finding the nearest prototype. A metric space is defined as the set of point embeddings together with a distance function defined on the set.
}
\label{fig_intro}
\end{figure*}

To overcome the data limitation, some researchers try to achieve 3D shape segmentation in an unsupervised or semi-supervised way. Early efforts \citep{golovinskiy2009consistent,hu2012co,huang2011joint,sidi2011unsupervised} tried to explore the underlying relevance among different shapes and perform face or patch primitive clustering in the descriptor space. Some recent works \citep{yi2018deep,wang2020few,yuan2020ross} approach the shape segmentation problem by transferring labels from a few template shapes to unseen shapes in a deformation-driven reconstruction process. Nevertheless, these two kinds of methods are trained class-by-class and cannot handle the segmentation of multiple classes as a whole. Another problem regarding these methods is that, by class-independent training, they cannot transfer the knowledge from learning one class (or multiple classes) to another.

To address these issues, in this paper, we introduce a novel meta-learning-based method for few-shot 3D shape segmentation. In the few-shot setting, we assume that adequate training samples are provided for some base classes and we are expected to generate a model that can perform accurate shape segmentation for novel classes with only a few labeled samples. Fig. \ref{fig_intro} illustrates the process of our meta-learning-based part segmentation model. Here we treat the part segmentation process for each shape class as a learning task. During meta-training, our model aims to learn a meta-metric learner that can adaptively learn a metric space for each sampled task. To achieve this, the meta-metric learner transforms input shapes from all classes into an embedding space where the class/task-specific metrics are defined. Then, for each class, a metric-based learner (i.e., segmentation module) is learned to dynamically modify distances between point embeddings thus maximizes in-part similarity while minimizing inter-part similarity. To achieve this, we average the point embeddings from the same semantic parts to get compact part-specific prototype representations and use a distance metric to encourage the intra-part similarity and inter-part dissimilarity. In this way, after meta-training, points from the same semantic part categories cluster around the associating prototype (points with red outlines in Fig. \ref{fig_intro}). While points from different part categories will have a larger distance in the metric space. In the meta-test stage, our metric-based learner (segmentation module) performs part segmentation for each point on the query shapes by assigning the label of its nearest prototype in the metric space.

The main contributions of our paper are summarized as follows:
\begin{itemize}
\item We introduce a novel meta-learning-based paradigm for 3D shape segmentation. Our method learns to learn a proper metric space for shape segmentation where points from the same part category lie close in the metric space. 
\item A metric-based learner is designed to learn well-separated part-specific prototypes from support shapes and performs segmentation for the query shape by matching each point to its nearest prototype.
\item A dual segmentation branch is proposed to make full use of the support information by switching support and query shapes and thereby encourages consistency between the support and query prototypes and improves the segmentation performance.
\item Our model achieves remarkable performance on  ShapeNet part dataset under the few-shot scenario, compared with well-established baselines and the state-of-the-art semi-supervised methods. We also demonstrated the transferring abilities of the proposed model on PSB dataset.
\end{itemize}

\section{Related Works}\label{sc_related}
\subsection{Supervised Shape Segmentation}
Inspired by the great success of CNN for image analysis, previous researchers try to process 3D point clouds by converting raw point clouds into 2D images or regular voxels. The first family of methods starts by rendering 3D point clouds into multiple 2D images and then uses traditional 2D CNNs for representative feature learning. Nevertheless, these methods tend to suffer from information loss \citep{klokov2017escape} and cannot be directly applied to other 3D tasks such as shape completion and large-scale semantic segmentation. Another family of methods directly partition 3D point clouds into volumetric grids and apply 3D CNNs on regular voxels \citep{zhou2018voxelnet,qi2016volumetric} to learn point features and per-point classification. However, the volumetric representation can cause memory and computation inefficiency. Recent researches~\citep{riegler2017octnet,tatarchenko2017octree,klokov2017escape} try to address this problem by Octree structure or KD-tree, but these methods suffer from performance degradation when the input point clouds are uneven sampled \citep{wu2019pointconv}. Another problem associated with volumetric-based methods is that they usually suffer from the loss of fine-grained detail information \citep{qi2016volumetric}.

The pioneering work PointNet \citep{qi2017pointnet} starts the trend of directly learning point signatures from raw 3D point sets. Although PointNet provides a simple and efficient architecture for point cloud signature learning and demonstrates satisfying performance for 3D shape segmentation, it lacks the ability to capture the local structures. PointNet++\citep{qi2017pointnet++} improves the method by using a hierarchy feature learning architecture and captures local structures at different scales. Subsequently, numerous variants, such as SO-Net \citep{li2018so}, PointCNN \citep{li2018pointcnn}, PointConv \citep{wu2019pointconv}, Kpconv \citep{thomas2019kpconv} and D-FCN \citep{wen2020directionally} have been proposed to improve the performance by carefully-designed point convolution operations. 

Another family of methods defines convolution layers on graph-structured data and is capable of capturing local topological structure. The defined graph convolutional networks apply multiple graph convolution layers to learn multi-scale node features over potentially long-range connections. For example, FeaStNet \citep{verma2018feastnet} proposed a graph-convolution operator that dynamically establishes the correspondences between kernel weights and local neighborhoods with arbitrary connectivity. GCN and its variants \citep{kipf2017semi,defferrard2016convolutional} defined graph-convolution layers based on graph spectral theory and graph Laplacian matrix. DGCNN \citep{wang2019dynamic} introduced a novel graph convolution operator named EdgeConv that can better capture local geometric features of point clouds while still maintaining permutation invariance. Nevertheless, the above supervised-learning-based methods mostly train deep neural networks in a fully supervised manner and thus require a huge number of labeled data to train the network parameters. In contrast, our proposed method is designed to solve the part segmentation problem in a few-shot scenarios and can lean accurate part segmentation with only a few labeled samples from novel classes.

\subsection{Semi-supervised Shape Segmentation}
Previous researches show that simultaneously segmenting a group of shape into consistent primitives, i.e., co-segmentation, tends to get better performance than segmenting each shape independently. Early efforts tried to extract hand-crafted descriptors for mesh faces or patch primitives and then perform simultaneous segmentation using clustering algorithms in the descriptor space. For example, Golovinskiy et al. proposed to treat shape segmentation as a graph clustering problem where each face is regarded as a node in the graph and corresponding faces in different meshes are connected to enable shape interaction \cite{golovinskiy2009consistent}. Following researches also explored shape co-segmentation using subspace clustering from hand-crafted descriptors in a pure unsupervised \citep{hu2012co,huang2011joint,sidi2011unsupervised} or semi-supervised \citep{wang2012active} manner. Researchers have also tried to introduce deep neural networks into the co-segmentation problem for representative feature learning. For example, Shu et al. \citep{shu2016unsupervised} adopted an auto-encoder network to transfer hand-crafted features to high-level features and then perform patch clustering in the embedding space. Chen et al. \citep{chen2019bae} introduced a branched auto-encoder network to approach the 3D shape co-segmentation problem that combines representation learning and subspace clustering in an end-to-end network.

Another family of methods tries to transfer labels from a few template shapes to unseen shapes in a deformation-driven reconstruction process \citep{yi2018deep,wang2020few,yuan2020ross}. Wang et al. introduced WPS-Net for few-shot point cloud segmentation \citep{wang2020few} . A coherent shape deformation network is designed to transfer labels from the template shapes to unlabeled shapes. Similarly, Yuan et al proposed a semi-supervised method for 3D mesh segmentation by shape deformation and label transferring \citep{yuan2020ross}. Nevertheless, these methods are trained class-by-class and cannot handle the segmentation of multiple classes as a whole. In contrast, our proposed method can simultaneously perform segmentation for shapes from multiple classes and also transfer the knowledge from existing learning tasks to current ones through a meta-learning mechanism.

\subsection{Meta Learning and Prototype Learning}
Meta-learning is a new learning paradigm that aims to develop a model that can quickly learn new concepts with a few training examples. It's one of the techniques widely used under the few-shot learning scenario. In general, existing meta-learning methods can be roughly divided into three categories, metric-based methods \citep{koch2015siamese,sung2018learning}, model-based methods \citep{santoro2016meta,munkhdalai2017meta}, and optimization-based methods \citep{ravi2016optimization,finn2017model}. The metric-based methods aim to learn a metric or distance function that can be used to measure the similarity between different data samples. The classification distribution can be calculated based on the similarity scores. The gradient-based methods focus on designing a learning mechanism specifically for fast adaption — a learning paradigm that can update the model parameters rapidly with only a few training steps. This fast adaption can be achieved by external memory storage and a RNN structure is commonly designed to storage previous information and rapidly incorporate new information \citep{santoro2016meta}. Others achieve fast parameter adaption by utilizing one neural network to predict the parameters of another neural network \citep{munkhdalai2017meta}. The optimization-based methods aim to develop a learning schema that a ``meta-learner'' is learned to efficiently update the parameters of the ``learner'' with a few samples and make it quickly adapt to new tasks. For example, Finn et al. proposed a method called Model-Agnostic Meta-Learning \citep{finn2017model} that aims to learn a model-agnostic parameter initialization for different tasks and the model can quickly update its parameters towards a specific task by a gradient-by-gradient technique.

In this paper, we mainly focused on the metric-based methods that try to learn an efficient distance metric for distinguishing samples from different classes. Commonly used metric-based methods include Siamese Neural Network \citep{koch2015siamese}, Matching Networks \citep{vinyals2016matching}, Relation Network \citep{sung2018learning}, and Prototypical Networks \citep{snell2017prototypical}. In Prototypical Networks \citep{snell2017prototypical}, the author proposed to characterize the property of each class with one representative feature vector (prototype) and matching each query sample to the nearest prototype. Inspired by this idea, we propose to learn part-specific prototypes and perform dense per-point 3D shape segmentation using the learned prototypes.

\section{Method}\label{sc_method}
In this section, we introduce the proposed method for Few-shot Part Segmentation (FPSeg). In section \ref{sc_pstate}, we states the few-shot part segmentation problem. We give a brief overview of the proposed method in section \ref{sc_overview}. The prototype learning and metric-based segmentation module are illustrated in section \ref{sc_proto} and section \ref{sc_similarity}. A dual segmentation module is further introduced to enhance the performance, which is presented in section \ref{sc_bilateral}. The training paradigm of our model is illustrated in section \ref{sc_pretrain}. 

\begin{figure*}
\begin{center}
\includegraphics[width=16cm,]{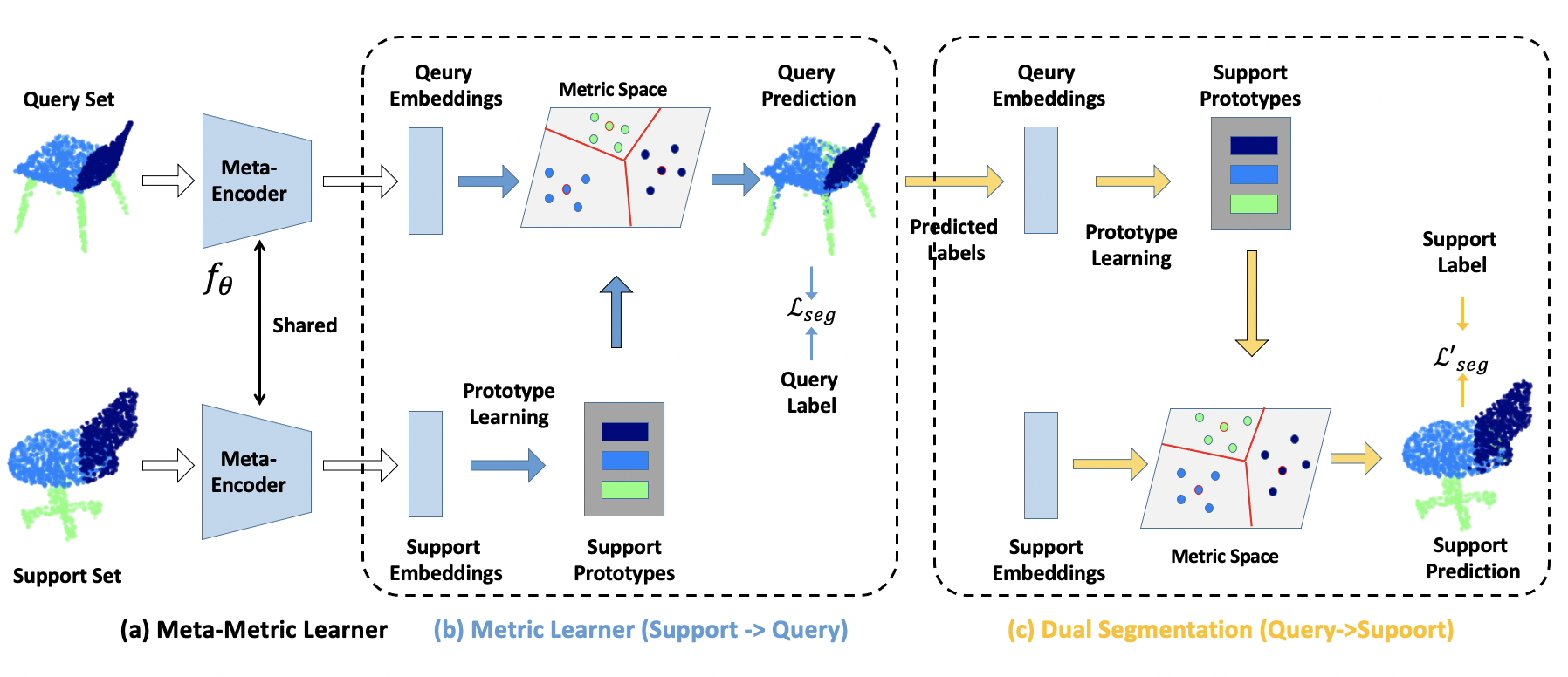}
\end{center}
\caption{Overview of the proposed few-shot part segmentation model. Our model consists of three parts. (a) The meta-metric learner transforms input shapes into an embedding space where the shape segmentation metrics are defined. (b) The metric learner formulates a class-specific segmentation metric using support information and performs per-point segmentation for the query shapes. In this block, a masked average pooling operation is applied to the support point embeddings to generate part prototypes using the ground truth part labels. After that, our model performs part labeling for each point in the query shapes by matching its feature embedding to the nearest prototype representation. Cosine distance is adopted as a distance function to measure the similarity between point embeddings and prototypes in the metric space. (c) The dual segmentation module performs query-to-support part segmentation. In this block, our model leverages the predicted part labels of query shapes to generate a new group of prototype representations (query prototypes) using a masked average pooling operation. Our model then performs part segmentation for the support shapes by labeling each point as the class of the nearest query prototype. Our model is optimized with metric learning-based losses for both support and query predictions ($\mathcal{L}_{seg}$ and $\mathcal{L}_{seg}'$ in the figure).}
\label{fig_overview}
\end{figure*}

\subsection{Problem Statement}\label{sc_pstate}
In this section, we introduce the proposed \textbf{\underline{F}}ew-shot \textbf{\underline{P}}art \textbf{\underline{Seg}}mentation (FPSeg) model. We first clarify the settings for the few-shot part segmentation problem. Suppose the given datasets are divided into two non-overlapping sets, $C_{s}$ and $C_{u}$. The problem of few-shot part segmentation aims at learning a part segmentation model from the dataset of base (seen) classes ($C_{s}$) that can perform part segmentation of 3D shapes from novel (unseen) classes ($C_{u}$) with only a few annotated samples. There is a large number of labeled 3D shapes from each of the seen classes, while the novel classes have only a few annotated 3D shapes. A few-shot part segmentation model should be able to learn meta-knowledge from the dataset of seen classes and obtain a segmentation model that can quickly adapt to the novel classes.

To achieve few-shot learning, a common solution is to build the part segmentation model using meta-learning techniques. Under the few-shot part segmentation scenario, a meta-learning algorithm learns to learn meta-knowledge from a large number of part segmentation tasks sampled from the seen classes and thus can generalize well to unseen categories. Each of the sampled tasks is called an episode and is designed to imitate the few-shot task by subsampling classes as well as shapes. The use of episode data makes the learning process more faithful to the test protocol and thereby enables good generalization with a few labels in the novel classes.

To facilitate model training and evaluation of a few-shot part segmentation model, we reorganize the training and testing data into several episodes. Each episode $\mathbf{E}_i$ is constructed from a set of support shapes $\mathbf{S}_i$ (with annotations) and a set of query shapes $\mathbf{Q}_i$. Given a K-shot part segmentation task, each support set $\mathbf{S}_i$ consists of $K$ annotated shapes per object category. We denote the support set as, $\mathbf{S}_i = \{(S_{i,k}, M_{ik})\}^K_{k=1}$ where $S_{i,k} \in \mathbb{R}^{N \times 3}$ denotes the input shapes, and $M_{i,k} \in \mathbb{R}^{N}$ denotes the part annotations. The query set $\mathbf{Q}_i$ contains $N_q$ shapes from the same set of object class $\mathcal{C}_i$ as the support set. The support shapes are used to generate part prototypes in the embedding space, and our model performs part segmentation for the query shapes via a distance metric. 

\subsection{Overview}\label{sc_overview}
Fig. \ref{fig_overview} gives an overview of the proposed few-shot part segmentation method. Given support and query shapes as inputs, our model starts with a shared meta-metric learner to extract point embeddings for both support and query shapes. We train a PointNet++ model on all base classes to get a powerful and robust backbone network and only keep these layers before the global max-pooling layer as the meta-encoder $f_{\theta}$. Then, the part labels of support shapes are applied to the point embeddings to generate part prototypes by a masked average pooling operation, as described in section \ref{sc_proto}. After that, our model predicts the part label for each point in the query shapes by matching its feature embedding to the nearest prototype representation. The task-specific metric space is optimized with a distance-based objective loss function and learns to maximize in-part similarity while minimizing inter-part similarity, as described in section \ref{sc_similarity}. A dual segmentation module that performs query-to-support part segmentation is further applied to encourage the learned prototypes to be consistent for support and query shapes, as described in section \ref{sc_bilateral}. 

\subsection{Prototype Learning}\label{sc_proto}

Our FPSeg model aims to learn well-separated prototype representation for each part category where a distance metric can be defined on. A part prototype can be regarded as a compressed feature vector that conveys the necessary information for distinguishing a specific part from others. In our method, the prototype representation is defined in the embedding space which ensures it to be robust towards different object categories and input variations. The hierarchy nature of PointNet++ encoder network also enables the learned prototypes to be robust towards different scales of input shapes.

To generate the part-specific prototype representations, we aggregate the deep features among different points of the same part category. More specifically, we leverage the part annotation masks of the supported shapes and apply a masked average pooling over the point embeddings to generate the prototypes for each part category. More specifically, given a part category $c$, the support set $\mathbf{S}_i=\{(S_{i,k}, M_{ik})\}^K_{k=1}$ and the extracted point embeddings $F_{i} = \{f_{i,k}\}_{k=1}^K$. The prototype representation of part $c$ can be calculated through a masked average pooling following:
\begin{equation}
P_c = \frac{1}{K} \sum_{k} \frac{\sum_{j} f_{i,k}^j \mathbb{I}(M_{i,k}^j=c)}{\sum_{j} \mathbb{I}(M_{i,k}^j=c)}
\label{eq_prototype}
\end{equation}
where $j$ denotes the point index in shape $S_{i,k}$, and $\mathbb{I}(\cdot)$ denotes an indicator function which outputs 1 if the input is true or 0 otherwise. Note that we average the prototypes among all $K$ support shapes to get the prototype representation for each part category.

\subsection{Metric Learner}\label{sc_similarity}
Given the part prototype representations, our metric learner module learns a metric space that measures the distance for each point embedding to the learned prototypes. Our model then predicts the part labels for each point in the query shapes by matching its corresponding point embedding to its nearest prototype in the embedding space. Specifically, we calculate the distance between the point embedding of each point in query shapes to all prototypes and search the nearest neighbor prototype and assign the corresponding class to this point. To enable back-propagation, we also calculate the part segmentation probabilities for each point by applying a softmax function over the distance values. Given the $j$th point $S_{i,q}^j$ in the $q$th query shape $S_{i,q}$, the part probability for class $c$ is calculated as:
\begin{equation}
H_{c}(S_{i,q}^j) = \frac{e^{-\alpha \mathcal{D}(f_{i,q}^j, P_c)}}{\sum_c e^{-\alpha \mathcal{D}(f_{i,q}^j, P_c)}}
\label{eq_prob}
\end{equation}
where $f_{i,q}^j$ denotes the point embedding for point $S_{i,q}^j$ and $\mathcal{D}(\cdot, \cdot)$ denotes the distance function defined in the embedding space, and $\alpha$ is a hyper-parameter to control probability distribution with regards to distance values. In our experiments, we set $\alpha$ to 2 since different values lead to similar performance. 

For the distance function $\mathcal{D}(\cdot, \cdot)$,  a common choice is to use either cosine distance or Euclidean distance. According to \citep{wang2019panet}, the cosine distance tends to be a better choice and can provide better performance for few-shot image segmentation. This is probably because cosine distance has bounded output and thus makes it easier to optimize. In this paper, we use cosine distance for distance function $\mathcal{D}(\cdot, \cdot)$ because it has bounded output and is easy to optimize. The final part segmentation map for $q$th query shape can be generated as:
\begin{equation}
M_{i,q}^j = \mathop{\arg \max}_c H_c(s_{i,q}^j)
\label{eq_pred}
\end{equation}
Given the predicted part probability map $H_{i,q}$ and the corresponding ground truth part label map $\tilde{M}_{i,q}$, we formulate the segmentation loss as:
\begin{equation}
\mathcal{L}_{seg} = -\frac{1}{N_q * N} \sum_{q} \sum_{j} log(H_{c}(S_{i,q}^j)) \mathbb{I}(\tilde{M}_{i,q}^j=c)
\label{eq_seg_loss}
\end{equation}
where $N_q$ and $N$ denote the number of query shapes in each episode and the number point in each query shape respectively. 

The above loss function will encourage the point embedding of a certain part category to be clustered around the corresponding part prototype meanwhile the clustering centers (i.e., prototypes) of different part categories are forced to be far from each other. After optimization, the learned prototypes are naturally well-separable for each part category.

\subsection{Dual Segmentation}\label{sc_bilateral}
To further boost the performance, we introduce a dual segmentation module to fully exploit the support information for few-shot learning. Our hypothesis is that if the model can accurately predict the part labels for query shapes from the prototypes extracted from support shapes, the prototypes extracted from query shapes should also be able to well segment support shapes. We, therefore, develop a bilateral segmentation module that takes query shapes and the predicted part labels as new `support' information and performs part labeling for the support shapes.

As illustrated in block (b) of Fig. \ref{fig_overview}, we generate query prototypes using point embeddings of query shapes and their predicted labels. The masked average pooling operation is also adopted here to generate these prototype representations. We perform part segmentation for the support shapes by labeling each point as the class of the nearest query prototype. The prototype learning and part labeling process is similar to as we described in section \ref{sc_proto}. The only difference here is that we are swapping the support and query shapes and use the predicted part labels of the query shapes as their ground truth labels in the masked average pooling layer.  We then formulate a prototype alignment loss by comparing the predicted part labels based on query prototypes and the ground truth part labels of support shapes. We add this prototype alignment regularization $\mathcal{L}_{seg}'$ to the final loss function as:
\begin{equation}
\mathcal{L} = \mathcal{L}_{seg} + \lambda \mathcal{L}_{seg}'
\label{eq_final_loss}
\end{equation}
where $\lambda$ denotes the hyper-parameter to balance these two loss terms. In our experiments, we set $\lambda$ to 1 because different values have a small influence on the final segmentation performance. By using this loss function, our model encourages a mutual alignment between the prototypes learned from support and query shapes. Our model can better use the support information in both support-to-query and query-to-support directions.

\subsection{Training Strategy}\label{sc_pretrain}
Our FPSeg model includes three training stages. The first stage is \textbf{segmentation pre-training}. Specifically, we train a conventional PointNet++ model on all base classes to get a powerful and robust meta-encoder network. In this stage, we use batch-wise training instead of episode data. Standard softmax cross-entropy loss is adopted to penalize the difference between predicted part labels and ground truth ones. The effect of the proposed pre-training process is validated by experiments in Section \ref{sc_exp_pre}. The second stage is \textbf{meta-training}. In this stage, our model performs meta-training on base classes using episode data. Our model is optimized with a metric learning-based objective loss function and learns to maximize in-part similarity while minimizing inter-part similarity. The third stage is \textbf{meta-finetuning}. In this stage, our model performs fine-tuning using a few labeled samples from the novel classes. We construct episodes for each novel class by randomly select $K/2$ labeled shapes as the support set and other shapes as the query set. In meta-finetuning, the network training process is the same as meta-training except for network inputs. Note that only a few labeled samples from novel classes are used for meta-finetuning. The effect of meta-finetuning process is validated by experiments in Section \ref{sc_exp_pre}.






\section{Experiments and Results}\label{sc_exp}

In this section, we conduct experiments to verify the effectiveness of the proposed modules in our FPSeg model and evaluate the part segmentation performance under the few-shot scenario. In section \ref{sc_exp_settings}, we introduce the experimental dataset and settings. We compare our method with both supervised methods and semi-supervised methods in section \ref{sc_exp_results}. We further show the performance of our FPSeg model with a different number of shots in section \ref{sc_exp_numshots}. We explore the effectiveness of pre-training and prototype alignment module in section \ref{sc_exp_pre} and section \ref{sc_exp_proto} respectively. The model size and inference time is given in section \ref{sc_exp_time}.

\subsection{Experimental Settings}\label{sc_exp_settings}
\noindent \textbf{Datasets. }We evaluate our few-shot part segmentation model (FPSeg) for part segmentation on ShapeNet part dataset \citep{yi2016scalable}. This dataset consists of 16,881 shapes from 16 object classes, with each object containing 2 to 6 parts categories, generating 50 part categories in total (with non-overlapping across object classes). Following \citep{qi2017pointnet++}, we randomly sample 2048 points from each object and use the official train/test split. To evaluate the part segmentation performance under few-shot settings, we evenly divide all 16 object classes into 4 folds, each with 4 object classes. We report the few-shot part segmentation performance on each fold, with the other three folds as the training set. 

\noindent \textbf{Evaluation Metrics.} We evaluate the performance of our FPSeg using intersection over union (IoU) following \citep{klokov2017escape}. For each shape instance, the IoU is defined as the average IoU of each part that belongs to the ground truth shape category. The mean IoU (mIoU) of each object class is calculated as an average over all the shapes in that class, and the overall average category mIoU (Cat. mIoU) is directly averaged over all classes.

\begin{table*}[t]
\begin{center}
\small
\caption{Part segmentation results on ShapeNet dataset. We report the performance when a model achieves the best average category mIoU on each novel set. Note that WPS-Net \citep{wang2020few} uses a 10-shot setting, while all other methods using 5-shot settings. Boldface and underline indicate the best and second  best performance.}
\label{table_1}
\resizebox{\textwidth}{!}{
\begin{tabular}{p{2.1cm}| *{4}{p{0.45cm}}| *{4}{p{0.47cm}}| *{4}{p{0.45cm}}| *{4}{p{0.47cm}} | p{0.5cm}}
\hline
& \multicolumn{4}{c|}{Novel Set 1} & \multicolumn{4}{c|}{Novel Set 2} & \multicolumn{4}{c|}{Novel Set 3}  & \multicolumn{4}{c|}{Novel Set 4} & \\
\hline
Method & aero & bag & cap & car & chair & earp. & guitar & knife & lamp & laptop & motor. & mug & pistol & rock. & skate. & table & Avg. \\
\hline
\#shapes & 2690 & 76 & 55 & 898 & 3758 & 69 & 787 & 392 & 1547 & 451 & 202 & 184 & 283 & 66 & 152 & 527 & \\
\hline
PointNet2+joint &61.9 & 64.3 & 82.9 & 41.1 & 84.6 & 66.6 & 70.6 & \underline{81.6} & 61.1 & \underline{94.9} & 25.9 &  48.6 & 62.2 & \underline{50.0} & \underline{64.8} & \underline{74.9} & 64.8\\
PointNet+ft & 48.7 & 45.0 & 72.4 & 24.2 & \underline{85.5} & 63.7 & 52.0 & 67.9 & 55.0 & 82.8 & 12.4 & 48.7 & 24.2 & 26.9 & 43.6 & 61.0 & 50.9\\
WPS-Net \citep{wang2020few} & \textbf{67.3} & \underline{74.4} & \textbf{86.3} & \underline{48.8} & 83.4 & \textbf{67.6} & 77.2 & 76.9 & \textbf{68.7} & 93.8 & \underline{43.0} & \underline{90.9} & \textbf{79.0} & 46.2 & 51.3 & 74.2 & 70.6\\
BAE-Net \citep{chen2019bae} & \underline{65.8} & 73.1 & 73.8 & - & 83.9 & 54.4 & \textbf{85.3} & 78.6 & 60.8 & 94.1 & \textbf{62.0} & \textbf{93.6} & \underline{76.7} & 41.8 & 64.0 & 72.5 & \underline{72.0} \\
\hline
FPSeg (Ours) & 62.2 & \textbf{80.3} & \underline{82.7} & \textbf{59.1} & \textbf{88.6} & \underline{66.2} & \underline{85.0} & \textbf{84.5} & \underline{61.7} & \textbf{95.3} & 38.4 & 90.3 & 76.6 & \textbf{62.9} & \textbf{72.3} & \textbf{75.9} & \textbf{73.9}\\
\hline
\end{tabular}}
\end{center}
\end{table*}

\subsection{Results Comparison}\label{sc_exp_results}
\noindent \textbf{Baselines.} We compare our FPSeg model with two competitive baselines. The first one is the vanilla PointNet++ model that we jointly train the model on all shapes from base classes and a few samples from novel classes. We term this baseline as PointNet2+joint. For the second baseline, we train PointNet++ model with the same pre-training stage as our FPSeg model, i.e., training on all shapes from base class; then we fine-tune the model with the same small set of samples from the novel classes. We term this baseline as PointNet2+ft. Comparing our method with these two baselines can help understand the advantage of the meta-learning-based method over traditional supervised methods. 

Table \ref{table_1} lists the performance of all comparing methods. From Table \ref{table_1} one can see that the two supervised PointNet++-based methods get a lot worse performance than the two few-shot methods (i.e., WPS-Net and our FPSeg). This reveals the generalization weakness of traditional supervised methods: without adequate training samples, supervised methods fail to properly transfer the knowledge learned from the base classes to novel classes. In contrast, by a proper design of the meta-learning framework, our FPSeg significantly enhances the performance by a large margin on novel classes. Specifically, our FPSeg model achieves an average category IoU that is 8.1\% better than the PointNet++ model using a joint training strategy. Besides, by comparing the performance of the two baseline methods, one can find that the PointNet++ model with a joint training strategy gets a lot better performance (+13.9\%) than the one with a fine-tuning strategy. 

We compare our FPSeg model with a recent few-shot part segmentation method, i.e., WPS-Net. WPS-Net relies on a coherent shape deformation network to transfer labels from template shapes to unseen shapes. For this method, we conduct model training and evaluation independently for each object class. From Table \ref{table_1} one can see that our FPSeg model achieves significantly better performance (+3.3\%) than the WPS-Net model. It should be noted that WPS-Net uses 10 exemplar shapes with annotations for each novel class, while our FPSeg model only uses 5 shots (labeled shapes) as the support set and can obtain better performance. 

We also compare our method with the state-of-the-art BAE-Net \citep{chen2019bae} method. For both BAE-Net and our FPSeg models, we report the average 5-shot performance by randomly select 5 examples and rerun the experiments 5 times. From Table \ref{table_1}, our FPSeg achieves significantly better performance (+1.9\%) than BAE-Net model. Specifically, our FPSeg model obtains a segmentation performance better than WPS-Net in 10 out of 15 object classes. It should be noted that the BAE-Net paper does not report the performance on the car category since it cannot work well on the car category. In contrast, our FPSeg can get reasonable performance on this category. If one removes the performance on the car category, our FPSeg can reach an even better performance of 74.8\%, which is 2.8\% better than the state-of-the-art BAE-Net model. Also, our model has two merits compared to BAE-Net: 1) our model is a lot faster than BAE-Net when new classes are given. This is because BAE-Net requires self-supervised training when new classes are given, while our FPSeg model only calls for a few fine-tuning steps on the new classes. 2) As mentioned in BAE-Net paper, their method can be sensitive to the initial parameters, where different runs may result in different segmentation results. In contrast, benefiting from meta-training, our method is quite stable compared to BAE-Net model. 

\begin{figure*}[t]
\centering
\includegraphics[width=16cm, height=8cm]{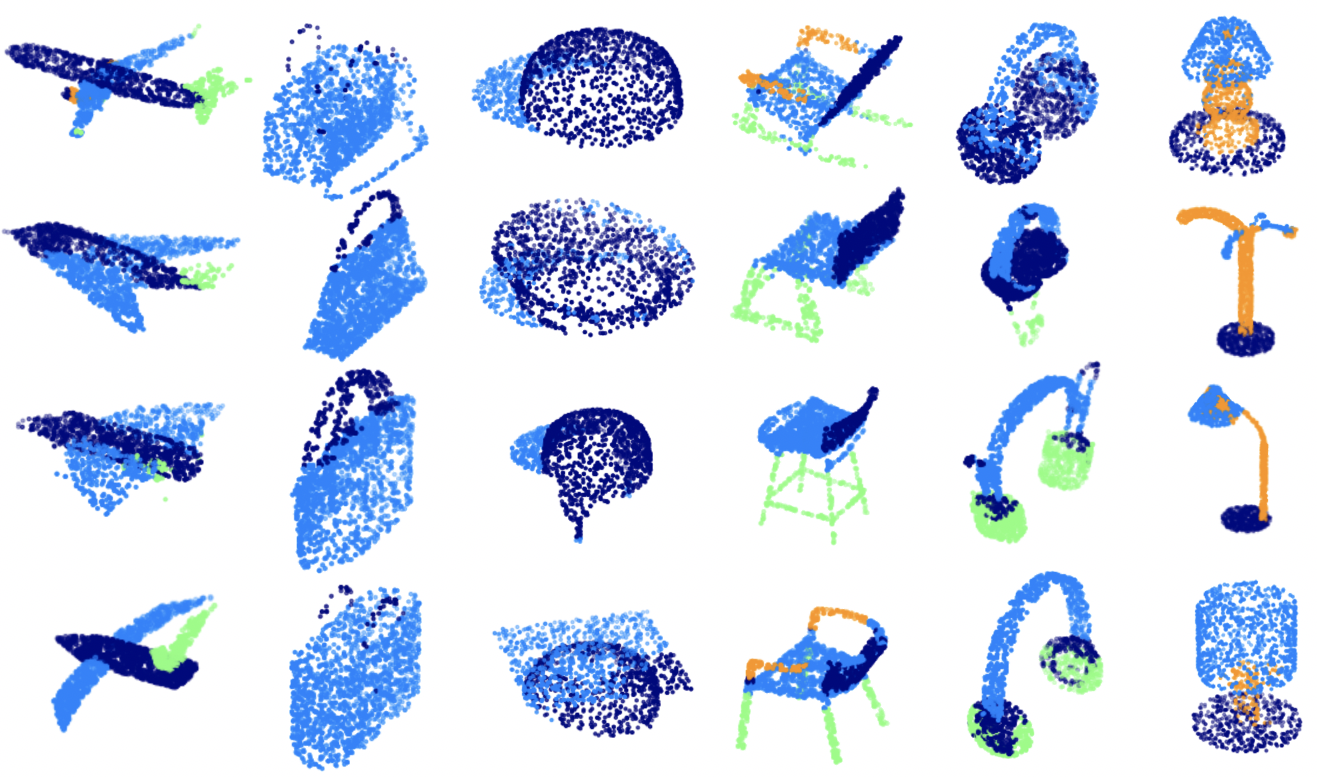}
\caption{Randomly selected examples of our few-shot part segmentation results.}
\label{exp_results}
\end{figure*}

\begin{figure}[h]
\centering
\includegraphics[width=16cm]{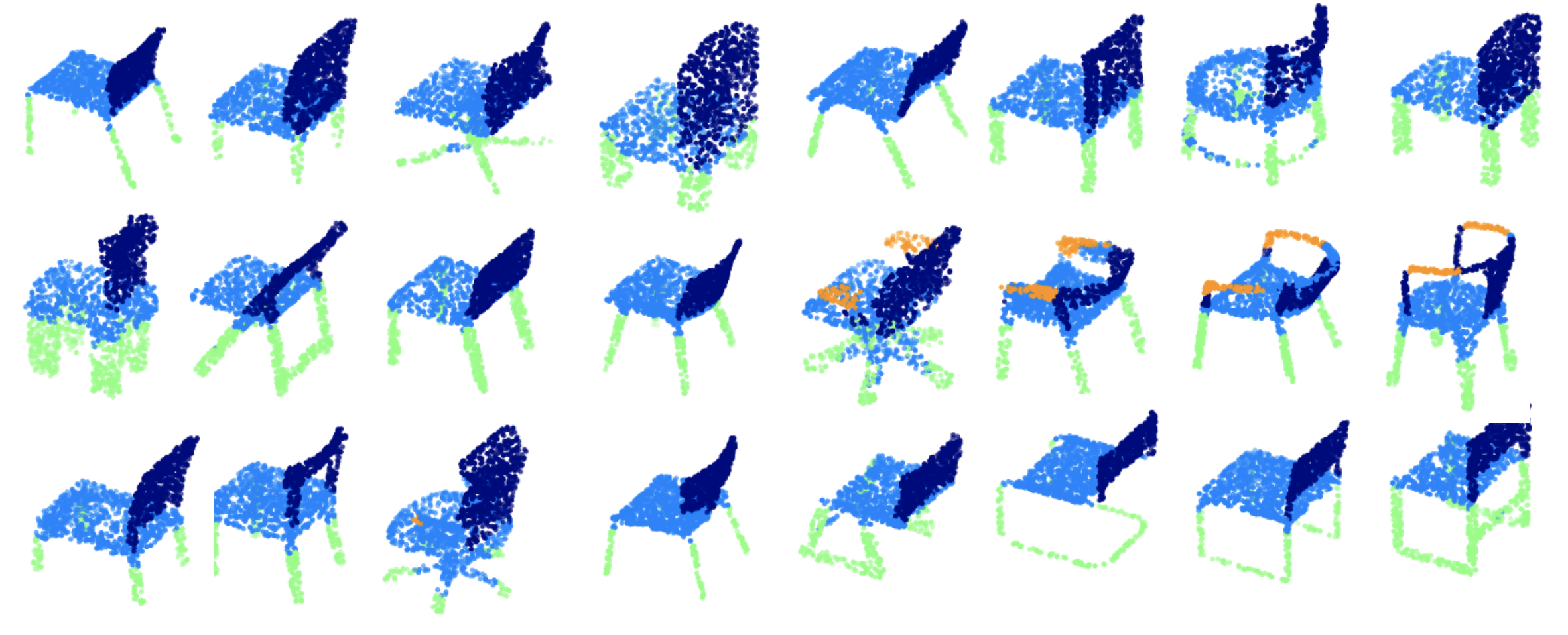}
\caption{Few-shot part segmentation results on ShapeNet chair category.}
\label{fig_chair}
\end{figure}

Fig. \ref{exp_results} shows randomly selected examples of our results for ShapeNet part segmentation. As shown in Fig. \ref{exp_results}, our FPSeg model can produce satisfying part segmentation results for shapes from different object classes and with different structures. More visualization results on the ShapeNet chair category are shown in Fig. \ref{fig_chair}.

\subsection{Different numbers of shots and sampled points}\label{sc_exp_numshots}
\noindent \textbf{Settings:} In this section, we carry our experiments to explore the performance of our FPSeg model using different numbers of support shapes and different numbers of sampled points on 3D shapes. For the first experiment, we randomly select 1, 2, 5, and 10 support shapes from the train-val split for each of the novel class and evaluate the performance of all shapes from the test split. For the second experiment, we fix the shot number to 5 and randomly sample 512, 1024, 2048, 4096 points for each shape. We conduct experiments on the novel set 2, including objects from four categories (chair, earphone, guitar, and knife).

\noindent \textbf{Results:} As shown in Table \ref{tab_num_shots}, with the increase of the number of support shapes, the performance increases quickly. Specifically, for the earphone category, using a 1-shot setting leads to a bad performance, while using 2-shot can significantly improve the performance. This is probably because using only 1 support shape cannot fully cover the main structural variations in the earphone category. While using 2 or more support shapes, our model can quickly learn to update the part prototypes and boost the performance. Moreover, from the bottom part of Table \ref{tab_num_shots}, one can see that our FPSeg model gets quite stable performance with different numbers of sampled points. Our FPSeg model achieves the best performance of 81.8\% mIoU on novel set 2, which is slightly better (0.7\%) than the performance using 2048 sampled points. In the following sections, 2048 points are sampled for each shape.

\begin{table}[h]
\centering
\caption{Performance on the novel set 2 using different numbers of shots and sampled points.}
\label{tab_num_shots}
\begin{tabular}{cccccc}
\hline
\#Shots & Chair & Earphone & Guitar & Knife & Avg.\\
\hline
1-shot & 85.6 & 16.6 & 80.0 & 81.1 & 65.8 \\
2-shot & 87.5 & 56.8 & 80.0 & 83.1 & 76.9 \\
5-shot & 88.6 & 66.2 & 85.0 & 84.5 & 81.1 \\
10-shot & 88.9 & 75.4 & 84.6 & 83.3 & 83.1\\
\hline
\hline
512-points & 87.9 & 64.0 & 86.2 & 83.6 & 80.4 \\
1024-points & 88.1 & 67.7 & 86.2 & 83.3 & 81.3\\
2048-points & 88.6 & 66.2 & 85.0 & 84.5 & 81.1 \\
4096-points & 88.4 & 68.4 & 86.7 & 83.5 & 81.8\\
\hline
\end{tabular}

\end{table} 

\subsection{Effect of Pre-training and Meta-finetuning}\label{sc_exp_pre}
To demonstrate the effect of segmentation pre-training and meta-finetuning process, we design three comparing methods: (a) our model without segmentation pre-training and meta-finetuning, (b) our model with segmentation pre-training but without meta-finetuning, and (c) our model with both segmentation pre-training and meta-finetuning. We list the performance on the novel set 2 (including chair, earphone, guitar, and knife) in Table \ref{tab_pretrain}. From Table \ref{tab_pretrain} one can see that our model without pre-training leads to significantly worse performance than the two models with pre-training. This indicates segmentation pre-training can potentially help to improve the transferability of our meta-learning model. Our finding is consistent with a recent work \citep{chen2020new} that uses classification pre-training for the few-shot classification task on 2D images. 

Moreover, by comparing models (b) and (c), we witness a significant performance boost obtained by meta-finetuning on novel classes. This is because both segmentation pre-training and meta-training are conducted only on base classes. Although our meta-training technique can improve transferability, there's still an inevitable gap if we directly apply the model trained on base classes to novel classes. By meta-finetuning on novel classes using a few annotated samples, our model learns to quickly adapt to novel classes and improves the performance by a large margin (+9.3\%). 

\begin{table}[h]
\centering
\caption{Effect of segmentation pre-training and meta-finetuning.}
\begin{tabular}{cccccc}
\hline
Method & Chair & earp. & Guitar & Knife & Avg.\\
\hline
Ours (a) &  67.9 & 46.6 & 73.2 & 51.4 & 59.8 \\
Ours (b) & 88.2 & 41.9 & 74.9 & 82.2 & 71.8\\
Ours (c) & 88.6 & 66.2 & 85.0 & 84.5 & 81.1\\
\hline
\end{tabular}

\label{tab_pretrain}
\end{table} 


\subsection{Effect of Prototype Alignment}\label{sc_exp_proto}
In this section, we conduct experiments to verify the effect of the proposed prototype alignment module for 3D shape segmentation. To achieve this, we carry experiments with and without the prototype alignment module. Results are listed in Table \ref{tab_align}. From Table \ref{tab_align} one can see that by adding the prototype alignment module, our model enjoys a performance boost of 1.3\% as indicated by the category average mIoU. 

\begin{table}[h]
\centering
\begin{tabular}{cccccc}
\hline
Method & Chair & earp. & Guitar & Knife & Avg.\\
\hline
Ours w/o align. & 87.2 & 65.1 & 83.6 & 83.1 & 79.8\\
Ours w/ align. & 88.6 & 66.2 & 85.0 & 84.5 & 81.1\\
\hline
\end{tabular}
\caption{Effect of prototype alignment.}
\label{tab_align}
\end{table} 


\subsection{Running Time and Model Capacity}\label{sc_exp_time}
We investigate the running time of our FPSeg model and the comparing methods including PointNet++ \cite{qi2017pointnet++}, WPS-Net \cite{wang2020few} and BAE-Net \cite{chen2019bae}. We calculate running time on the chair category with 704 test shapes. Experiments are conducted on a single TITAN Xp with a batch size of 8. Note that WPS-Net and BAE-Net require self-supervised training to tune its parameters from scratch for each unseen class. In our experiments, we follow the setting of WPS-Net and BAE-Net papers to train WPS-Net for 5000 steps and BAE-Net for 200000 steps. In contrast, our FPSeg model only calls for a few fine-tuning steps on the new classes, we report the training time when our FPSeg model is finetuned for 50 steps.

As shown in Table \ref{tab_time}, our FPSeg model has a similar model capacity as the PointNet++ model, which is smaller than the two self-supervised methods, i.e., WPS-Net and BAE-Net.
For the training time, WPS-Net and BAE-Net require 14m52s and 2h46m for self-supervised training, while our FPSeg model only takes 80s for a few fine-tuning steps. This demonstrates that our FPSeg model can be trained a lot faster than WPS-Net and BAE-Net models when new test classes are given.
We also report the per batch inference time in Table \ref{tab_time}. As can be seen in Table \ref{tab_time}, our FPSeg model takes a slightly longer inference time than other comparing models. This is because, in our FPSeg model, the part prototypes need to be calculated using the support shapes. Specifically, it takes 267ms for our FPSeg model to predict the part labels for each batch of input shapes when the batch size set is 8.

\begin{table}[h]
\centering
\begin{tabular}{cccc}
\hline
Method & \#Parameters & Training Time & Inference Time\\
\hline
PointNet++ & 1.4M & 15s & 254ms\\
WPS-Net & 2.6M & 14m52s & 91ms\\
BAE-Net & 5.2M & 2h46m8s & 168ms \\
FPSeg (ours) & 1.7M  & 80s &  267ms\\
\hline
\end{tabular}
\caption{Running time and model capacity.}
\label{tab_time}
\end{table}

\subsection{Transfer to Mesh Segmentation}
To further demonstrate the generalization ability of our proposed FPSeg model for 3D shape segmentation, we conduct segmentation experiments on PSB dataset \citep{chen2009benchmark} using our model trained on ShapeNet part dataset. We follow the settings of \citep{xu2017directionally} to generate training/test split and choose one example per-class from the training split as the support set and all shapes in the test split for model evaluation. Four categories (glasses, octopus, table, and teddy) from the PSB dataset are chosen as the novel classes and we conduct meta-finetuning on one support shape from each class. We compare our method with DCN \citep{xu2017directionally}, which is one of the state-of-the-art supervised methods for mesh segmentation, and ROSS \citep{yuan2020ross} that performs one-shot mesh segmentation using a deformation network-based method.

\begin{figure*}[]
\centering
\includegraphics[width=14cm, height=6cm]{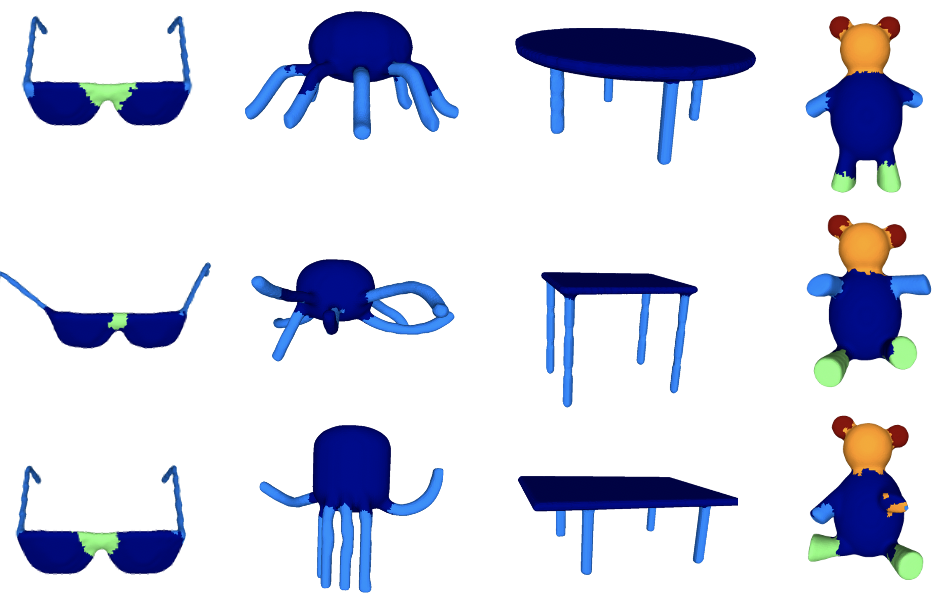}
\caption{Segmentation results on PSB dataset.}
\label{exp_psb}
\end{figure*}

Table \ref{tab_psb} lists the segmentation performance as indicated by overall accuracy for each object class. As shown in Table \ref{tab_psb}, our FPSeg model gets a performance better than the few-shot-based method ROSS and even better than the STOA supervised method DCN. Specifically, our FPSeg model obtains an average accuracy of 93.1\% for the selected classes, which demonstrates a powerful generalization ability of our FPSeg model for 3D shape segmentation. It should be noted that both DCN and ROSS use a more powerful backbone network while our FPSeg model uses the PointNet++ backbone for feature extraction. Moreover, DCN uses a post-processing step to refine the segmentation results while our FPSeg model achieves better performance without any post-processing. Fig. \ref{exp_psb} shows randomly selected examples of our segmentation results on the PSB dataset.

\begin{table*}[]
\centering
\begin{tabular}{cccccc}
\hline
Method & Glasses & Octopus & Table & Teddy & Avg.\\
\hline
DCN \citep{xu2017directionally} & 87.4 & 90.7 & 93.9 & 92.7 & 91.1 \\
ROOS \citep{yuan2020ross}  & \textbf{90.7}  & \textbf{91.7} & 95.2& 93.7 & 92.8 \\
FPSeg (ours) & 90.6 & 88.6 & \textbf{99.1} & \textbf{94.0} & \textbf{93.1} \\
\hline
\end{tabular}
\caption{Segmentation performance on PSB dataset. Note that DCN uses fully supervised training, while ROOS and our FPSeg use a 1-shot setting.}
\label{tab_psb}
\end{table*} 


\section{Conclusion}
We introduce a meta-learning-based method for few-shot 3D shape segmentation. Our method learns to learn part-specific prototype representations from the support shapes in the embedding space and dynamically modify distances between point embeddings thus maximizes in-part similarity. It performs per-point segmentation over the query shapes by matching each point to its nearest prototype. A prototype alignment branch is applied to perform segmentation in an opposite direction that can make better use of the support information and implicitly encourages consistency between the support and query prototypes. We demonstrate the superior performance of the proposed on the ShapeNet part dataset under the few-shot scenario, compared with well-established baseline methods and the state-of-the-art semi-supervised methods.



\bibliographystyle{spbasic}
\bibliography{egbib}

\begin{thebibliography}{45}
\providecommand{\natexlab}[1]{#1}
\providecommand{\url}[1]{{#1}}
\providecommand{\urlprefix}{URL }
\expandafter\ifx\csname urlstyle\endcsname\relax
  \providecommand{\doi}[1]{DOI~\discretionary{}{}{}#1}\else
  \providecommand{\doi}{DOI~\discretionary{}{}{}\begingroup
  \urlstyle{rm}\Url}\fi
\providecommand{\eprint}[2][]{\url{#2}}

\bibitem[{Chen et~al.(2009)Chen, Golovinskiy, and
  Funkhouser}]{chen2009benchmark}
Chen X, Golovinskiy A, Funkhouser T (2009) A benchmark for 3d mesh
  segmentation. Acm transactions on graphics (tog) 28(3):1--12

\bibitem[{Chen et~al.(2020)Chen, Wang, Liu, Xu, and Darrell}]{chen2020new}
Chen Y, Wang X, Liu Z, Xu H, Darrell T (2020) A new meta-baseline for few-shot
  learning. arXiv preprint arXiv:200304390

\bibitem[{Chen et~al.(2019)Chen, Yin, Fisher, Chaudhuri, and
  Zhang}]{chen2019bae}
Chen Z, Yin K, Fisher M, Chaudhuri S, Zhang H (2019) Bae-net: Branched
  autoencoder for shape co-segmentation. In: Proceedings of the IEEE
  International Conference on Computer Vision, pp 8490--8499

\bibitem[{Defferrard et~al.(2016)Defferrard, Bresson, and
  Vandergheynst}]{defferrard2016convolutional}
Defferrard M, Bresson X, Vandergheynst P (2016) Convolutional neural networks
  on graphs with fast localized spectral filtering. In: NIPS

\bibitem[{Finn et~al.(2017)Finn, Abbeel, and Levine}]{finn2017model}
Finn C, Abbeel P, Levine S (2017) Model-agnostic meta-learning for fast
  adaptation of deep networks. arXiv preprint arXiv:170303400

\bibitem[{Golovinskiy and Funkhouser(2009)}]{golovinskiy2009consistent}
Golovinskiy A, Funkhouser T (2009) Consistent segmentation of 3d models.
  Computers \& Graphics 33(3):262--269

\bibitem[{Hu et~al.(2012)Hu, Fan, and Liu}]{hu2012co}
Hu R, Fan L, Liu L (2012) Co-segmentation of 3d shapes via subspace clustering.
  In: Computer graphics forum, Wiley Online Library, vol~31, pp 1703--1713

\bibitem[{Huang et~al.(2011)Huang, Koltun, and Guibas}]{huang2011joint}
Huang Q, Koltun V, Guibas L (2011) Joint shape segmentation with linear
  programming. In: ACM transactions on graphics (TOG), ACM, vol~30, p 125

\bibitem[{Jiang et~al.(2018)Jiang, Wu, and Lu}]{jiang2018pointsift}
Jiang M, Wu Y, Lu C (2018) Pointsift: A sift-like network module for 3d point
  cloud semantic segmentation. arXiv preprint arXiv:180700652

\bibitem[{Kipf and Welling(2017)}]{kipf2017semi}
Kipf TN, Welling M (2017) Semi-supervised classification with graph
  convolutional networks. ICLR

\bibitem[{Klokov and Lempitsky(2017)}]{klokov2017escape}
Klokov R, Lempitsky V (2017) Escape from cells: Deep kd-networks for the
  recognition of 3d point cloud models. In: Computer Vision (ICCV), 2017 IEEE
  International Conference on, IEEE, pp 863--872

\bibitem[{Koch et~al.(2015)Koch, Zemel, and Salakhutdinov}]{koch2015siamese}
Koch G, Zemel R, Salakhutdinov R (2015) Siamese neural networks for one-shot
  image recognition. In: ICML deep learning workshop, Lille, vol~2

\bibitem[{Li et~al.(2018{\natexlab{a}})Li, Chen, and Hee~Lee}]{li2018so}
Li J, Chen BM, Hee~Lee G (2018{\natexlab{a}}) So-net: Self-organizing network
  for point cloud analysis. In: Proceedings of the IEEE conference on computer
  vision and pattern recognition, pp 9397--9406

\bibitem[{Li et~al.(2018{\natexlab{b}})Li, Bu, Sun, Wu, Di, and
  Chen}]{li2018pointcnn}
Li Y, Bu R, Sun M, Wu W, Di X, Chen B (2018{\natexlab{b}}) Pointcnn:
  Convolution on x-transformed points. In: Advances in Neural Information
  Processing Systems, pp 820--830

\bibitem[{Litany et~al.(2017)Litany, Remez, Rodola, Bronstein, and
  Bronstein}]{litany2017deep}
Litany O, Remez T, Rodola E, Bronstein A, Bronstein M (2017) Deep functional
  maps: Structured prediction for dense shape correspondence. In: Proceedings
  of the IEEE International Conference on Computer Vision, pp 5659--5667

\bibitem[{Munkhdalai and Yu(2017)}]{munkhdalai2017meta}
Munkhdalai T, Yu H (2017) Meta networks. Proceedings of machine learning
  research 70:2554

\bibitem[{Qi et~al.(2016)Qi, Su, Nie{\ss}ner, Dai, Yan, and
  Guibas}]{qi2016volumetric}
Qi CR, Su H, Nie{\ss}ner M, Dai A, Yan M, Guibas LJ (2016) Volumetric and
  multi-view cnns for object classification on 3d data. In: Proceedings of the
  IEEE conference on computer vision and pattern recognition, pp 5648--5656

\bibitem[{Qi et~al.(2017{\natexlab{a}})Qi, Su, Mo, and Guibas}]{qi2017pointnet}
Qi CR, Su H, Mo K, Guibas LJ (2017{\natexlab{a}}) Pointnet: Deep learning on
  point sets for 3d classification and segmentation. Proc Computer Vision and
  Pattern Recognition (CVPR), IEEE 1(2):4

\bibitem[{Qi et~al.(2017{\natexlab{b}})Qi, Yi, Su, and
  Guibas}]{qi2017pointnet++}
Qi CR, Yi L, Su H, Guibas LJ (2017{\natexlab{b}}) Pointnet++: Deep hierarchical
  feature learning on point sets in a metric space. In: Advances in Neural
  Information Processing Systems, pp 5099--5108

\bibitem[{Ravi and Larochelle(2016)}]{ravi2016optimization}
Ravi S, Larochelle H (2016) Optimization as a model for few-shot learning

\bibitem[{Riegler et~al.(2017)Riegler, Osman~Ulusoy, and
  Geiger}]{riegler2017octnet}
Riegler G, Osman~Ulusoy A, Geiger A (2017) Octnet: Learning deep 3d
  representations at high resolutions. In: Proceedings of the IEEE Conference
  on Computer Vision and Pattern Recognition, pp 3577--3586

\bibitem[{Santoro et~al.(2016)Santoro, Bartunov, Botvinick, Wierstra, and
  Lillicrap}]{santoro2016meta}
Santoro A, Bartunov S, Botvinick M, Wierstra D, Lillicrap T (2016)
  Meta-learning with memory-augmented neural networks. In: International
  conference on machine learning, pp 1842--1850

\bibitem[{Shu et~al.(2016)Shu, Qi, Xin, Hu, Wang, Zhang, and
  Liu}]{shu2016unsupervised}
Shu Z, Qi C, Xin S, Hu C, Wang L, Zhang Y, Liu L (2016) Unsupervised 3d shape
  segmentation and co-segmentation via deep learning. Computer Aided Geometric
  Design 43:39--52

\bibitem[{Sidi et~al.(2011)Sidi, van Kaick, Kleiman, Zhang, and
  Cohen-Or}]{sidi2011unsupervised}
Sidi O, van Kaick O, Kleiman Y, Zhang H, Cohen-Or D (2011) Unsupervised
  co-segmentation of a set of shapes via descriptor-space spectral clustering,
  vol~30. ACM

\bibitem[{Snell et~al.(2017)Snell, Swersky, and Zemel}]{snell2017prototypical}
Snell J, Swersky K, Zemel R (2017) Prototypical networks for few-shot learning.
  In: Advances in neural information processing systems, pp 4077--4087

\bibitem[{Su et~al.(2015)Su, Maji, Kalogerakis, and
  Learned-Miller}]{su2015multi}
Su H, Maji S, Kalogerakis E, Learned-Miller E (2015) Multi-view convolutional
  neural networks for 3d shape recognition. In: Proceedings of the IEEE
  international conference on computer vision, pp 945--953

\bibitem[{Su et~al.(2018)Su, Jampani, Sun, Maji, Kalogerakis, Yang, and
  Kautz}]{su2018splatnet}
Su H, Jampani V, Sun D, Maji S, Kalogerakis E, Yang MH, Kautz J (2018)
  Splatnet: Sparse lattice networks for point cloud processing. In: Proceedings
  of the IEEE Conference on Computer Vision and Pattern Recognition, pp
  2530--2539

\bibitem[{Sung et~al.(2018)Sung, Yang, Zhang, Xiang, Torr, and
  Hospedales}]{sung2018learning}
Sung F, Yang Y, Zhang L, Xiang T, Torr PH, Hospedales TM (2018) Learning to
  compare: Relation network for few-shot learning. In: Proceedings of the IEEE
  Conference on Computer Vision and Pattern Recognition, pp 1199--1208

\bibitem[{Tatarchenko et~al.(2017)Tatarchenko, Dosovitskiy, and
  Brox}]{tatarchenko2017octree}
Tatarchenko M, Dosovitskiy A, Brox T (2017) Octree generating networks:
  Efficient convolutional architectures for high-resolution 3d outputs. In:
  Proc. of the IEEE International Conf. on Computer Vision (ICCV), vol~2, p~8

\bibitem[{Thomas et~al.(2019)Thomas, Qi, Deschaud, Marcotegui, Goulette, and
  Guibas}]{thomas2019kpconv}
Thomas H, Qi CR, Deschaud JE, Marcotegui B, Goulette F, Guibas LJ (2019)
  Kpconv: Flexible and deformable convolution for point clouds. In: Proceedings
  of the IEEE International Conference on Computer Vision, pp 6411--6420

\bibitem[{Verma et~al.(2018)Verma, Boyer, and Verbeek}]{verma2018feastnet}
Verma N, Boyer E, Verbeek J (2018) Feastnet: Feature-steered graph convolutions
  for 3d shape analysis. In: Proceedings of the IEEE conference on computer
  vision and pattern recognition, pp 2598--2606

\bibitem[{Vinyals et~al.(2016)Vinyals, Blundell, Lillicrap, Wierstra
  et~al.}]{vinyals2016matching}
Vinyals O, Blundell C, Lillicrap T, Wierstra D, et~al. (2016) Matching networks
  for one shot learning. In: Advances in neural information processing systems,
  pp 3630--3638

\bibitem[{Wang et~al.(2017)Wang, Pelillo, and Siddiqi}]{wang2017dominant}
Wang C, Pelillo M, Siddiqi K (2017) Dominant set clustering and pooling for
  multi-view 3d object recognition. In: Proceedings of British Machine Vision
  Conference (BMVC), vol~12

\bibitem[{Wang et~al.(2019{\natexlab{a}})Wang, Liew, Zou, Zhou, and
  Feng}]{wang2019panet}
Wang K, Liew JH, Zou Y, Zhou D, Feng J (2019{\natexlab{a}}) Panet: Few-shot
  image semantic segmentation with prototype alignment. In: Proceedings of the
  IEEE International Conference on Computer Vision, pp 9197--9206

\bibitem[{Wang et~al.(2020)Wang, Li, and Fang}]{wang2020few}
Wang L, Li X, Fang Y (2020) Few-shot learning of part-specific probability
  space for 3d shape segmentation. In: Proceedings of the IEEE/CVF Conference
  on Computer Vision and Pattern Recognition, pp 4504--4513

\bibitem[{Wang and Solomon(2019)}]{wang2019deep}
Wang Y, Solomon JM (2019) Deep closest point: Learning representations for
  point cloud registration. In: Proceedings of the IEEE International
  Conference on Computer Vision, pp 3523--3532

\bibitem[{Wang et~al.(2012)Wang, Asafi, Van~Kaick, Zhang, Cohen-Or, and
  Chen}]{wang2012active}
Wang Y, Asafi S, Van~Kaick O, Zhang H, Cohen-Or D, Chen B (2012) Active
  co-analysis of a set of shapes. ACM Transactions on Graphics (TOG) 31(6):165

\bibitem[{Wang et~al.(2019{\natexlab{b}})Wang, Sun, Liu, Sarma, Bronstein, and
  Solomon}]{wang2019dynamic}
Wang Y, Sun Y, Liu Z, Sarma SE, Bronstein MM, Solomon JM (2019{\natexlab{b}})
  Dynamic graph cnn for learning on point clouds. Acm Transactions On Graphics
  (tog) 38(5):1--12

\bibitem[{Wen et~al.(2020)Wen, Yang, Li, Peng, and Chi}]{wen2020directionally}
Wen C, Yang L, Li X, Peng L, Chi T (2020) Directionally constrained fully
  convolutional neural network for airborne lidar point cloud classification.
  ISPRS Journal of Photogrammetry and Remote Sensing 162:50--62

\bibitem[{Wu et~al.(2019)Wu, Qi, and Fuxin}]{wu2019pointconv}
Wu W, Qi Z, Fuxin L (2019) Pointconv: Deep convolutional networks on 3d point
  clouds. In: Proceedings of the IEEE Conference on Computer Vision and Pattern
  Recognition, pp 9621--9630

\bibitem[{Xu et~al.(2017)Xu, Dong, and Zhong}]{xu2017directionally}
Xu H, Dong M, Zhong Z (2017) Directionally convolutional networks for 3d shape
  segmentation. In: Proceedings of the IEEE International Conference on
  Computer Vision, pp 2698--2707

\bibitem[{Yi et~al.(2016)Yi, Kim, Ceylan, Shen, Yan, Su, Lu, Huang, Sheffer,
  and Guibas}]{yi2016scalable}
Yi L, Kim VG, Ceylan D, Shen IC, Yan M, Su H, Lu C, Huang Q, Sheffer A, Guibas
  L (2016) A scalable active framework for region annotation in 3d shape
  collections. ACM Transactions on Graphics (ToG) 35(6):1--12

\bibitem[{Yi et~al.(2018)Yi, Huang, Liu, Kalogerakis, Su, and
  Guibas}]{yi2018deep}
Yi L, Huang H, Liu D, Kalogerakis E, Su H, Guibas L (2018) Deep part induction
  from articulated object pairs. arXiv preprint arXiv:180907417

\bibitem[{Yuan and Fang(2020)}]{yuan2020ross}
Yuan S, Fang Y (2020) Ross: Robust learning of one-shot 3d shape segmentation.
  In: The IEEE Winter Conference on Applications of Computer Vision, pp
  1961--1969

\bibitem[{Zhou and Tuzel(2018)}]{zhou2018voxelnet}
Zhou Y, Tuzel O (2018) Voxelnet: End-to-end learning for point cloud based 3d
  object detection. In: Proceedings of the IEEE Conference on Computer Vision
  and Pattern Recognition, pp 4490--4499

\end{thebibliography}

\end{document}